\documentclass[journal, 10pt]{IEEEtran}

\IEEEoverridecommandlockouts 

\usepackage{mathtools}
\usepackage{amssymb}
\usepackage{algorithmic}
\usepackage{array}
\usepackage{graphicx}
\usepackage{booktabs}
\usepackage{multirow}
\usepackage{dcolumn}
\usepackage[para, online, flushleft]{threeparttable}
\usepackage{tablefootnote}
\usepackage{accents}
\usepackage{textcomp}
\usepackage{hyperref}
\usepackage{gensymb}
\usepackage{upgreek}
\usepackage{mathptmx}
\usepackage{tikz}
\usepackage{balance}

\newcolumntype{L}[1]{>{\raggedright\arraybackslash}p{#1}}
\newcolumntype{C}[1]{>{\centering\arraybackslash}p{#1}}
\newcolumntype{R}[1]{>{\raggedleft\arraybackslash}p{#1}}
\title{Design and Control of a Compact Series Elastic Actuator Module for Robots in MRI Scanners}
\author{Binghan~He, Naichen~Zhao, David~Y.~Guo, Charles~H.~Paxson, Alfredo~De~Goyeneche, Michael~Lustig, Chunlei~Liu, and Ronald~S.~Fearing
\thanks{ This work was supported by the National Institutes of Health under Grant R01MH127104. 
The content is solely the responsibility of the authors and does not necessarily represent the official views of the National Institutes of Health. 
\emph{(Corresponding author: Binghan He.)}} 
\thanks{ Binghan He was with the Department of Electrical Engineering and Computer Sciences, University of California, Berkeley, CA 94720 USA. He is now with the Department of Mechanical, Aerospace, and Industrial Engineering, The University of Texas at San Antonio, San Antonio, TX 78249 USA (e-mail: binghan.he@utsa.edu).} 
\thanks{ Naichen Zhao, Alfredo De Goyeneche, Michael Lustig, Chunlei Liu, and Ronald S. Fearing are with the Department of Electrical Engineering and Computer Sciences, University of California, Berkeley, CA 94720 USA (e-mail: nzhao11235@berkeley.edu, asdegoyeneche@berkeley.edu, mikilustig@berkeley.edu, chunlei.liu@berkeley.edu, ronf@berkeley.edu).} 
\thanks{ David Y. Guo and Charles H. Paxson are with the Department of Mechanical Engineering, University of California, Berkeley, CA 94720 USA (e-mail: dguo001@berkeley.edu, chpaxson@berkeley.edu).}}

\newcommand\copyrighttext{\footnotesize\sf This article has been accepted for publication in IEEE/ASME Transactions on Mechatronics. \copyright \ 2026 IEEE. DOI: 10.1109/TMECH.2026.3658501}

\newcommand\copyrightnotice{\begin{tikzpicture}[remember picture,overlay] \node[anchor=north, xshift=0pt] at (current page.north) {\fbox{\parbox{\dimexpr\textwidth-\fboxsep-\fboxrule\relax}{\copyrighttext}}}; \end{tikzpicture} \vspace{-10pt}}

\begin{document}

\maketitle

\copyrightnotice

\begin{abstract}
Robotic assistance has broadened the capabilities of magnetic resonance imaging (MRI)-guided medical interventions, yet force-controlled actuators tailored for MRI environments remain limited. 
In this study, we present a novel MRI-compatible rotary series elastic actuator (SEA) module that employs velocity-sourced ultrasonic motors for force-controlled operation within MRI scanners. 
Unlike prior MRI-compatible SEA designs, our module uses a transmission force sensing SEA architecture, with four off-the-shelf compression springs placed between the gearbox and motor housings. 
To enable precise torque control, we develop a controller based on a disturbance observer, specifically designed for velocity-sourced motors. 
This controller improves torque regulation, even under varying external impedance, enhancing the actuator’s suitability for MRI-guided medical interventions. 
Experimental validation confirms effective torque control in both 3 Tesla MRI and non-MRI settings, achieving a 5\% settling time of 0.05 seconds and steady-state error within 2.5\% of the actuator’s maximum output torque. 
Notably, the controller maintains consistent performance across both low and high impedance conditions. 
\end{abstract}


\section{Introduction} \label{sec:introduction}
\IEEEPARstart{W}{hile} robots have proven effective in enhancing the precision and time efficiency of MRI-guided interventions across various medical applications \cite{su2022state}, safety remains a formidable challenge for robots operating within MRI environments due to several factors. 
As the robots assume full control of medical procedures, the reliability of their operation becomes paramount. 
Precise control over robot forces is particularly crucial to ensure safe interaction within the MRI environment. 
Furthermore, the confined space in the MRI bore complicates the safe operation of human-robot interaction, presenting challenges to maneuverability.
However, there exists a notable scarcity of force-controlled robot actuators specifically tailored for MRI applications. 

Unlike non-MRI environments, the strong magnetic field within MRI scanners renders conventional electromagnetic robot actuators unsafe for medical applications in MRI. 
While fluidic actuators offer MRI compatibility \cite{guo2018compact, dong2019high}, the lack of back-drivability in hydraulic actuators poses significant risks in medical scenarios involving human interaction. 
Pneumatic actuators \cite{yu2008comparison}, known for their compliance, enhance safety in human-robot interactions, yet their restricted force output and controllability confine their applicability to specific medical procedures, such as needle-based interventions \cite{monfaredi2018mri, xiao2020mr}.
Recent advances in continuous \cite{gunderman2023modeling, pan2024cornerstone} and stepper \cite{farimani2018introducing, groenhuis2018rapid} pneumatic motors have enhanced torque output and displacement precision, expanding their potential applications.
Hydrostatic transmissions, driven by remote electromagnetic motors, offer another solution for MRI-compatible actuation by transferring motor forces through extended hoses \cite{gassert2006mri}. 
However, the high piston seal friction inherent in hydrostatic transmissions introduces additional uncertainty in force control precision.

Non-magnetic electric motors, such as electrostatic motors \cite{yamamoto2005evaluation} and ultrasonic motors (USMs) \cite{fischer2008mri, krieger2011development}, have been developed to provide both precision and compatibility with MRI environments. 
However, electrostatic motors require an extremely high voltage source to generate significant output force and are not commercially available.
Akin to hydraulic actuators, USMs encounter inherent challenges such as high impedance and lack of backdrivability. 
Typically, USMs are utilized as velocity sources, restricting their application to the problem of controlling human-robot interaction forces.

In \cite{pratt1995series}, the series elastic actuator (SEA) concept introduces a unique architecture for robot actuators, enhancing safety during human-robot interaction. 
By placing a spring element in series with a rigid actuator, compliance is added. While traditional SEAs use force-sourced motors, this approach also benefits velocity-sourced actuators \cite{wyeth2008demonstrating}. 
The spring element functions as a force sensor via Hooke's law, enabling the control of actuator force to be translated into the control of spring deflection.

The development of MRI-compatible SEAs is initially presented in \cite{sergi2015interaction} for the implementation of an MRI-guided rehabilitation robot. 
This system achieves force-controlled actuation by integrating an USM, a pair of extension springs serving as the spring element, and a linear bearing. 
The evolution of this technology continued with a rotary SEA in \cite{senturk2018mri}. 
However, this particular actuator relies on an electromagnetic motor located outside the MRI room, connected via a lengthy cable-driven transmission, thus limiting its applicability for in-bore MRI robotic tasks. 
Notably, while compact SEA designs based on electromagnetic motors have been extensively investigated in non-MRI robot applications \cite{kong2011compact, paine2013design, lee2016configuration, kim2021compact}, the development of integrated compact SEAs utilizing MRI-compatible motors remains an area yet to be fully explored.

Achieving effective force control for velocity-sourced SEAs poses another significant challenge, especially in MRI applications. 
For the MRI-compatible SEAs in \cite{sergi2015interaction, senturk2018mri}, proportional (P) or proportional-integral (PI) controllers have been used for actuator force control. 
However, their performance and stability heavily depend on the external impedance affecting the actuator. 
In scenarios where MRI robot applications involve substantial variation in external impedance, the closed-loop system behavior becomes uncertain. 
To address this challenge, the disturbance observer (DOB) method \cite{chen2015disturbance} has been integrated into numerous SEA control applications. 
The DOB calculates the difference between the input of the actual SEA system and that of a reference model. 
This difference is then compensated to ensure that the adjusted system behavior matches the reference model. 
Since the parameters of the DOB are independent of external impedance variations, it enhances the robustness of force control.
Although this approach has been studied for SEAs employing force-sourced or position-controlled motors \cite{kong2011compact, paine2013design, oh2016high, nakamura2021torque}, its application to velocity-sourced motors has not been thoroughly examined.

While this study focuses on velocity-sourced SEAs for MRI robot applications, advanced adaptive control strategies have also been explored for force-sourced SEAs.
For instance, adaptive control techniques such as model reference adaptive control (MRAC) \cite{losey2016time} and reinforcement learning-based tuning \cite{sambhus2023real} have demonstrated promising tracking performance.
However, these approaches lack formal stability guarantees under time-varying dynamics, raising concerns in safety-critical applications such as MRI-guided interventions, where abrupt human contact may occur.
Adaptive robust control methods \cite{calanca2018understanding} address this limitation by combining MRAC with sliding mode control, offering enhanced robustness.
In \cite{yan2018generalized}, a model predictive control method was applied to SEA position control, incorporating online estimation of nonparametric uncertainty, which is conceptually similar to DOB-based methods.

\begin{figure} [!tbp]
\centering
\includegraphics[width=0.99\linewidth]{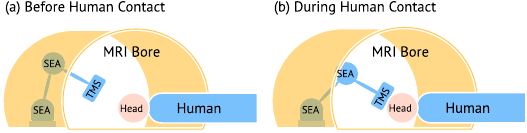}
\caption{
In (a) and (b), we show a conceptual illustration of a potential MRI-guided brain stimulation application that motivates our SEA module design, where an SEA robot maneuvers a transcranial magnetic stimulator (TMS) toward a patient's head.
The SEA torque controller needs to address the challenges posed by both (a) the minimal external impedance of the medical device before it makes contact with the patient and (b) the substantial impedance encountered upon contact with the human body.
}
\label{fig:tms}
\end{figure}

In this study, we aim to address the above-mentioned challenges associated with MRI-compatible force-controlled actuators for robotic applications within MRI scanners. 
Diverging from prior MRI-compatible SEAs developed for rehabilitation robots \cite{sergi2015interaction, senturk2018mri}, which consistently interact with human patients, the SEA in this study is motivated by MRI-guided brain stimulation applications illustrated in Fig.~\ref{fig:tms}.
This requires a compact design for in-bore actuation and robust force control to handle varying external impedance, ensuring the robot can safely control forces during the transition from free motion to contact.

\begin{figure} [!tbp]
\centering
\includegraphics[width=0.99\linewidth]{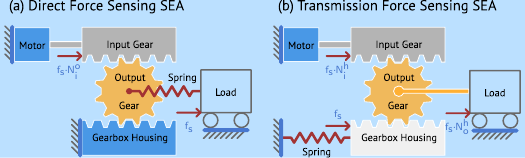}
\caption{
In (a), a conventional direct force sensing SEA positions its spring element between the output gear and the load. 
Contrarily, in (b), a transmission force sensing SEA situates its spring element between the gearbox housing and the ground. 
The transmission force sensing SEA architecture enables the construction of a more compact SEA suitable for MRI scanners. 
$\mathsf{f_s}$ denotes the spring force. 
$\mathsf{N_{i}^{o}}$, $\mathsf{N_{i}^{h}}$, and $\mathsf{N_{o}^{h}}$ denote the gear ratios between the input and the output, between the input and the gearbox housing, and between the output and the gearbox housing.
}
\label{fig:TFSEA}
\end{figure}

To this end, we focus on the design and torque control of a 1-DOF SEA module, establishing it as a foundational joint unit for future multi-DOF MRI-compatible robots.
The primary contributions of this paper are summarized as follows.
\begin{itemize}
\item[(1)] 
In Sec.~\ref{sec:design}, we introduce a compact MRI-compatible SEA module that seamlessly integrates an USM, a rotary gearbox, and a spring element into a transmission force sensing SEA structure, as depicted in Fig.~\ref{fig:TFSEA}. 
In contrast to previous MRI-compatible SEAs \cite{sergi2015interaction, senturk2018mri}, our design is the first rotary SEA that operates entirely within an MRI scanner. 
\item[(2)] 
In Sec.~\ref{sec:control}, we present a DOB-based controller for actuator torque control in our compact SEA module.
This controller distinguishes itself from previous DOB-based SEA control strategies \cite{kong2011compact, paine2013design, nakamura2021torque}, which were primarily designed for force-sourced or position-controlled motors.
Notably, our DOB-based controller is specifically crafted for SEA systems featuring velocity-sourced motors, such as the MRI-compatible velocity-controlled USM integrated into our SEA module.
\item[(3)] 
In Sec.~\ref{sec:experiment}, we demonstrate the consistent performance of our SEA module in both 3~Tesla MRI and non-MRI environments. 
In comparison to direct force control strategies employed in velocity-sourced SEAs~\cite{wyeth2008demonstrating, sergi2015interaction, senturk2018mri}, our DOB-based controller exhibits its capability to handle both low-impedance conditions, representing no-contact scenarios, and high-impedance conditions due to rigid environmental contact without requiring changes in controller structure or tuning.
\end{itemize}

\begin{figure*} [!tbp]
\centering
\includegraphics[width=0.99\linewidth]{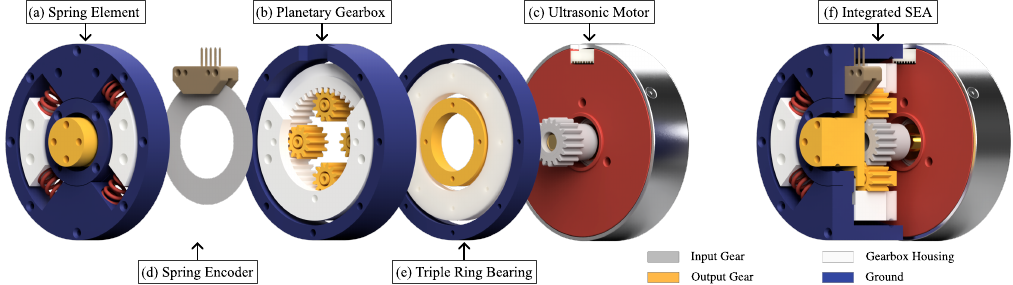}
\caption{
The architecture of our compact SEA module comprises essential components: a spring element (a), a gearbox (b), an USM (c), and an optical spring encoder (d). 
Integration is facilitated by a triple ring bearing (e), allowing the entire framework to fit seamlessly into a cylindrical space, as depicted in (f), with a diameter of $80$ mm and a total length of $66$ mm.}
\label{fig:design}
\end{figure*}

\begin{table*} [!tbp]
\caption{Comparison to Other SEAs for MRI and Non-MRI Applications}
\label{tab:cSEA}
\centering
\includegraphics[width=1.00\linewidth]{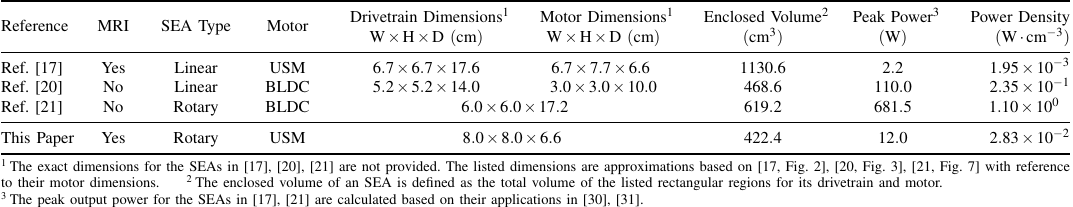}
\end{table*}

\section{Mechanical Design} \label{sec:design}

In this section, we present a compact SEA design that seamlessly integrates an USM, a rotary gearbox, and a spring element into a unified module. 
Towards the conclusion, we offer a comparative analysis with previous MRI-compatible SEAs and assess the MRI-compatibility of our SEA module.

For a multi-DOF robot system, each SEA module must be scaled appropriately based on payload, joint location, and robot kinematics. 
In this paper, we focus on the wrist joint because it exhibits the lowest effective output impedance under no-contact conditions, making it a representative scenario for validating the robustness of our torque controller across extreme impedance variations between no-contact and contact, as discussed in Sec.~\ref{sec:experiment}.

Specifically, we consider an output torque of at least $2.0~\mathrm{N \cdot m}$ and a rated velocity of at least $18 \degree \cdot \mathrm{s}^{-1}$. 
These specifications correspond to a nominal SEA module that (i) drives a $1 \sim 3~\mathrm{kg}$ commercial TMS coil rotating about a wrist joint with a radius of $5~\mathrm{cm}$, (ii) moves the TMS from the forehead to the back of the head (a $180^\circ$ rotation at the wrist joint) in $10~\mathrm{sec}$, and (iii) applies an additional $5 \sim 10~\mathrm{N}$ contact force. 
Given the slow motion considered here, the effect of Lorentz forces on the TMS coil is less significant than the effects of gravity and contact forces.

\subsection{Transmission Force Sensing SEA} \label{sec:TFSEA}

While a conventional SEA has the output of a geared motor attached to the load through a spring element and the gearbox housing rigidly attached to the ground \cite{pratt1995series}, our design adopts the configuration in \cite{lee2016configuration}. 
In this configuration the spring element is placed between the gearbox housing and the ground, and the output gear is rigidly attached to the load.
In this paper, we distinguish the conventional SEA as a direct force sensing series elastic actuator (FSEA) and the variant introduced in \cite{lee2016configuration} as a transmission force sensing series elastic actuator (TFSEA), which is also referred to as a differential elastic actuator in \cite{shu2024impact} or as a compact planetary elastic actuator in \cite{lee2016configuration} when a planetary gearbox is used.

In Fig.~\ref{fig:TFSEA}, the structural difference between an FSEA and a TFSEA is illustrated. 
The TFSEA presents several advantages for implementation in MRI scanners. 
Notably, the spring element is fixed to the ground and motor housing, rather than moving with the actuator output, allowing for compact integration of the motor, gearbox, and spring element. 
Moreover, placing the electronic components of the spring deflection encoder on the ground side eliminates the need for encoder cable movement. 

In Fig.~\ref{fig:design}(b), our SEA incorporates a planetary gearbox design. 
This configuration, akin to the TFSEA illustrated in Fig.~\ref{fig:TFSEA}, features a sun gear as the input gear, an assembly of four planet gears serving as the output gear, and an inner ring gear encompassing the planet gears to form the gearbox housing.
The gear ratio $\mathsf{N_{i}^{o}}$ between the input and output is chosen within a range of $\frac{1}{5} \sim \frac{1}{3}$ to avoid excessively small sun or planet gears, which are difficult to manufacture due to material strength limitations. 
In our prototype, the gear ratio is established at $\mathsf{N_{i}^{o}} = \frac{3}{10}$.

\subsection{Spring Element} \label{sec:spring}

In the realm of rotary SEAs, the adoption of helical torsion springs as elastic elements is well-explored, as demonstrated in \cite{kong2011compact, lee2016configuration}. 
Specifically, the spring selection in \cite{lee2016configuration} plays a pivotal role in achieving a compact integration of the SEA by encompassing a brushless direct current (BLDC) motor with a helical torsion spring. 
However, while implementing a helical torsion spring simplifies rotary SEA design, it does not enhance compactness when paired with a disk-shaped USM, which typically has a larger outer diameter and shorter length compared to the BLDC motor used in \cite{lee2016configuration}. 
To address this, a TFSEA design introduced in \cite{kim2021compact} employs a customized planar torsion spring element to enhance overall compactness. 
Nonetheless, this planar torsion spring requires more metals than a helical spring, potentially leading to more interference in MR imaging processes. 

Fig.~\ref{fig:design}(a) presents the spring element design, drawing inspiration from the SEA in \cite{wyeth2008demonstrating}. 
To address the previously highlighted concerns, the torsion spring element features four off-the-shelf 316 stainless steel (SS 316) helical compression springs arranged in parallel. 
The loading of these four parallel springs is achieved through two sliders affixed to the housing of the planetary gearbox. 
Each helical compression spring, with a spring constant of $4.3$ $\mathrm{N \cdot mm ^ {-1}}$, places its centerline $25$ mm from the rotation axis under no load.

When torque is applied, the torsion spring element allows for a rotational range of $\pm 12 \degree$, shifting the centerline of each helical compression spring within a range of $24.14 \sim 25.48$ mm from the rotation axis of the torsion spring element. 
Despite this, the total torsion spring rate remains relatively constant, ranging from $10.48 \sim 10.65$ $\mathrm{N \cdot m \cdot rad^{-1}}$. 
This consistency is due to the opposing shifts of the centerlines of the two springs on either side of the slider, effectively balancing each other out. 
At full compression of $12 \degree$, the spring element generates maximum torques of $2.2$ $\mathrm{N \cdot m}$ at the gearbox housing and $3.2$ $\mathrm{N \cdot m}$ at the output.

\subsection{Compact Integration} \label{sec:integration}

Fig.~\ref{fig:design}(f) depicts a half-section view of our compact SEA integration, incorporating the spring element, the planetary gearbox, and a WLG-75-R USM (Tekceleo, Mougins, France).
At the peak SEA output power of $12.0\ \mathrm{W}$, the USM delivers a rated velocity of $4.8\ \mathrm{rad \cdot s^{-1}}$ and an output torque of $2.5\ \mathrm{N \cdot m}$.
The maximum achievable output torque, limited by the torsion spring element, is $3.2\ \mathrm{N \cdot m}$.
This SEA configuration delivers sufficient output torque and velocity to satisfy the wrist-joint requirements defined at the beginning of Sec.~\ref{sec:design}.

To assemble the integrated SEA and mitigate off-axis movement of each component with respect to the ground, we have designed a triple-ring ball bearing. 
As illustrated in Fig.~\ref{fig:design}(e), this triple-ring bearing secures its inner ring to the output gear, the middle ring to the gearbox housing, and the outer ring to the ground. 
Within the spring element, an additional ball bearing enforces the constraint between the ground and the output gear. 
To measure the spring deflection, we employ an EM2-2-10000-I rotary optical encoder module (US Digital, Vancouver, WA, USA), with the encoder attached to the grounded housing in the spring element and the codewheel affixed to the gearbox housing. 
While most SEA components were fabricated using computer numerical control (CNC) machining with a tolerance of $\pm 0.1$~mm, the gearbox was customized for this prototype using multi-jet fusion (MJF) $3$D printing.

As a whole, the spring element, planetary gearbox, and bearing seamlessly fit within a cylindrical space with an outer diameter of $80$ mm and a total length of $39.5$ mm. 
This comprises a $12$ mm length for the spring element, $18$ mm for the gearbox, and $9.5$ mm for the bearing. 
When combined with a $22.5$ mm thick USM and a $4$ mm thick USM mounting plate, the integrated SEA has a total length of $66$ mm.

\begin{figure} [!tbp]
\centering
\includegraphics[width=0.99\linewidth]{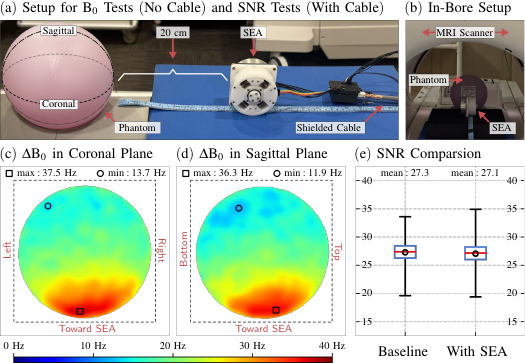}
\caption{
In the setup for the image artifact tests, the SEA is positioned 20 cm in front of a spherical phantom in a 3T GE MR750W scanner (GE Healthcare, Waukesha, WI, USA), as shown in (a). 
For the $\mathrm{B_0}$ tests, the setup is placed inside the scanner without the cable in (b). For the SNR tests, a shielded cable connects the SEA to the control room while the SEA remains unpowered during imaging. 
The resulting $\mathrm{B_0}$ field deviation ($\Delta \mathrm{B_0}$) caused by the SEA is shown in the coronal (c) and sagittal (d) planes. No shimming was performed between scans. 
The dashed lines in (c) and (d) indicate the boundaries of the imaging field of view (FOV). 
The SNR comparison between the baseline (no SEA) and the SEA-present setting is shown in (e), with each distribution computed within a thresholded phantom region: the circle marks the mean SNR, the red line indicates the median, the box edges mark the 25th and 75th percentiles, and the whiskers show the full range.
}
\label{fig:image}
\end{figure}

\begin{table} [!tbp]
\caption{Materials Used in Spring Element, Gearbox, and Bearings} \label{tab:material}
\centering
\includegraphics[width=1.00\linewidth]{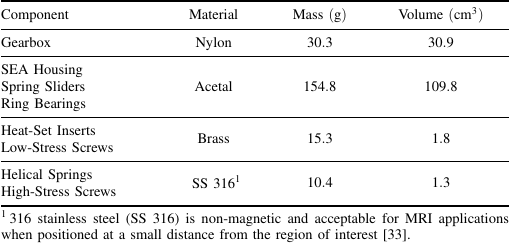}
\end{table}

In Table~\ref{tab:cSEA}, we present a dimensional comparison between our integrated SEA module and other SEAs employing USMs and electromagnetic motors. 
Our SEA module achieves a power-to-volume density comparable to the linear MRI-compatible SEA described in \cite{sergi2015interaction}. 
While our SEA module has a lower power-to-volume density than non-MRI-compatible SEAs using electromagnetic motors \cite{paine2013design, lee2016configuration}, this is expected due to the inherently lower power density of commercial USMs compared to electromagnetic motors. 
However, the advantage of using a USM in our design is also clear from prior work \cite{sergi2015interaction}, as it provides inherently MR-safe actuation that conventional electromagnetic motors cannot offer.

\subsection{MRI Compatibility} \label{sec:image}

As discussed in \cite{sergi2015interaction}, the MRI compatibility of an SEA includes the safety and actuator functionality in MRI environment and non-interference in the imaging processes. 
The safety in MRI environment is ensured by incorporating MRI-safe motor, encoder, and materials into the SEA module. 
In Sec.~\ref{sec:T-A}, we will show the torque control performance of our SEA in an MRI scanner. 

The extent of interference in MR imaging is largely determined by the magnetic susceptibility and conductivity of materials \cite{elhawary2008case}.
Similar to the SEA in \cite{sergi2015interaction}, the spring element, gearbox, and bearings of our SEA module are fabricated from non-magnetic and non-conductive polymers, with only a small volume of metal used in the helical compression springs and fasteners, as detailed in Table~\ref{tab:material}.
However, the primary source of imaging interference remains the commercial USM, which raises greater concerns for image artifacts than the pneumatic motors used in \cite{gunderman2023modeling, pan2024cornerstone, farimani2018introducing, groenhuis2018rapid}.
Therefore, as recommended in \cite{fischer2008mri, krieger2011development}, USMs should be positioned $20 \sim 30$ cm away from the region of interest during MR imaging.

In Fig.~\ref{fig:image}, the SEA is positioned 20 cm from a spherical phantom, which has an outer diameter of 18 cm. 
The main magnetic field ($\mathrm{B_0}$) deviation induced by the SEA remains within the range of $11.9 \sim 37.5$ Hz ($0.28 \sim 0.88 \ \upmu \mathrm{T}$), which is acceptable for the 3T MRI scanner used in this study and can be further corrected with MRI $\mathrm{B_0}$ field shimming techniques including using existing shimming coils or custom local shimming coils \cite{kim2002regularized, stockmann2018vivo}. 
The average signal-to-noise ratio (SNR) in the phantom volume changes less than $1\%$ between the baseline setting without the SEA and with the unpowered SEA present ($27.3 \rightarrow 27.1$).
In the appendix, we provide a detailed explanation of the MR image artifact tests.

\section{Modeling and Control} \label{sec:control}

In this section, we develop a torque controller for velocity-sourced SEAs, ensuring force regulation in MRI robotic applications with both low and high external impedance, as shown in Fig.~\ref{fig:tms}. 
To enhance clarity, all variables introduced in our control design are referenced in the output frame of our SEA. First, we define $\uptheta_e$ as the SEA's output displacement. 
Accordingly, $\uptheta_a$, $\uptheta_s$, and $k_s$ denote the reflected USM's displacement, spring deflection, and spring stiffness respectively, transitioning from their original reference frames to the output frame. 
The output torque $\uptau_s$ exerted by the spring in our SEA is computed from the spring encoder measurements of $\uptheta_s$ using Hooke's law: $\uptau_s = k_s \cdot \uptheta_s$. 
The output displacement $\uptheta_e$, is computed as the difference between the USM's encoder measurements of $\uptheta_a$ and the spring encoder measurements of $\uptheta_s$. 
In addition, we define $\upomega_e$, $\upomega_a$, and $\upomega_s$ as the time derivative of $\uptheta_e$, $\uptheta_a$, and $\uptheta_s$.

\subsection{Modeling}

\begin{figure} [!tbp]
\centering
\includegraphics[width=0.99\linewidth]{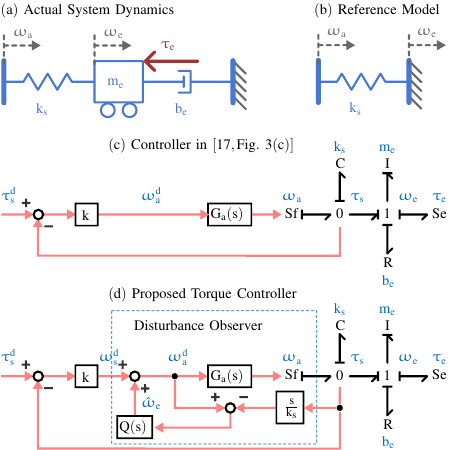}
\caption{
In (a), the velocity-sourced SEA encounters uncertain and finite external impedance, posing challenges to torque control. 
Conversely, our DOB's reference model in (b) has an infinitely large external impedance attached to its output. 
Unlike the direct torque controller depicted in (c), as proposed in \cite[Fig.~3(c)]{sergi2015interaction}, our control framework in (d) incorporates the DOB to compel the actual velocity-sourced SEA system to emulate the behavior of the reference model. 
This framework ensures that a proportional controller can achieve effective force reference tracking despite the uncertain external impedance. 
The bond graphs in (c) and (d) represent the SEA model in (a). 
In the bond graphs, $\mathrm{I}$, $\mathrm{R}$, $\mathrm{C}$, $\mathrm{Sf}$, and $\mathrm{Se}$ denote the nodes representing inertia, damper, spring, velocity source, and force source, respectively, in a mechanical system.
}
\label{fig:model}
\end{figure}

In Fig.~\ref{fig:model}(a), we model the external system dynamics that attach to the output of our SEA as a mass $m_e$, a damper $b_e$, and an external force source $\uptau_e$. 
Let $Z_s(s)$ and $Z_e(s)$ be the impedance models for the actuator spring and the external system, respectively. 
According to the equations of motion with $\uptheta_a$ and $\uptheta_e$, we define $Z_s(s)$ and $Z_e(s)$ as
\begin{align}
& Z_s (s) \triangleq \frac{\uptau_s}{\upomega_s} = \frac{\uptau_s}{\upomega_a - \upomega_e} = k_s s ^ {- 1}, \label{eq:Zs} \\
& Z_e (s) \triangleq \frac{\uptau_s - \uptau_e}{\upomega_e} = m_e s + b_e. \label{eq:Ze}
\end{align} 
Combining \eqref{eq:Zs} and \eqref{eq:Ze}, we eliminate $\upomega_e$ in the expression and derive an expression for $\uptau_s$ in terms of the motor velocity input $\upomega_a$ and the external force input $\uptau_e$:
\begin{align}
\uptau_s 
& = G_{sa} (s) \cdot \upomega_a + G_{se} (s) \cdot \uptau_e, \label{eq:fs}
\end{align}
with the transfer functions $G_{sa} (s)$ and $G_{se} (s)$ define as
\begin{align}
& G_{sa} (s) \triangleq \frac{Z_s Z_e}{Z_s + Z_e} = \frac{k_s (m_e s + b_e)}{m_e s ^ 2 + b_e s + k_s}, \label{eq:Gsa} \\
& G_{se} (s) \triangleq \frac{Z_s    }{Z_s + Z_e} = \frac{k_s}{m_e s ^ 2 + b_e s + k_s}. \label{eq:Gse}
\end{align} 
By including a velocity-controlled USM transfer function $G_a (s) \triangleq \frac{\upomega_a}{\upomega_a^d}$ with a $50 \ \mathrm{Hz}$ bandwidth, we obtain
\begin{align}
\uptau_s 
& = G_{sa}^d (s) \cdot \upomega_a^d + G_{se} (s) \cdot \uptau_e,
\end{align}
where $\upomega_a^d$ is the desired motor velocity and $G_{sa}^d (s) \triangleq G_{sa} (s) G_a (s)$. 
In addition, $\upomega_a^d$ is set within $\pm 5.2 \ \mathrm{rad \cdot s ^ {-1}}$ due to the velocity limits of the USM.

\subsection{DOB-Based Torque Control} \label{sec:fs}

In Fig.~\ref{fig:tms}, we illustrate how the SEA torque controller must manage both minimal external impedance before patient contact and significant impedance upon contact. 
According to \eqref{eq:fs}, the SEA torque controller is more effective with larger external impedance, resulting in a higher gain $G_{sa}$ for the motor input $\upomega_a$ and a lower gain $G_{se}$ for the disturbance force $\uptau_e$. 
Particularly, as $Z_e \rightarrow \infty$ in \eqref{eq:Gsa} and \eqref{eq:Gse}, we have $G_{sa} \rightarrow Z_s$ and $G_{se} \rightarrow 0$, leading to our DOB's reference model shown in Fig.~\ref{fig:model}(b).
This reference model $G_{sa}(s) = Z_s (s)$, an integrator multiplied by actuator spring constant $k_s$, enables stabilization with a single P controller $\upomega_a^d = k \cdot (\uptau_s^d - \uptau_s)$ for effective force reference tracking. 

In Fig.~\ref{fig:model}(d), we show the proposed DOB for imposing the reference model without actually having $Z_e = \infty$. Under this DOB, we estimate the output velocity $\upomega_e$ as
\begin{align}
\hat{\upomega}_e = Q (s) \cdot (\upomega_a ^ d - Z_s ^ {-1} \cdot \uptau_s) \label{eq:ve}
\end{align}
where $Q(s)$ is a low-pass filter of sufficient order to ensure the observer is causal.
By implementing 
\begin{align}
\upomega_a ^ d & = \upomega_s ^ d + \hat{\upomega}_e, \label{eq:va-d}
\end{align}
we have the desired velocity $\upomega_s^d$ for spring deflection as the new control input of the actuator system. Substituting \eqref{eq:ve} and \eqref{eq:va-d} into \eqref{eq:fs}, we obtain 
\begin{align}
\uptau_s 
& = G_{ss}^d (s) \cdot \upomega_s^d + \hat{G}_{se} (s) \cdot \uptau_e. \label{eq:fs-2}
\end{align}
where 
\begin{align}
& G_{ss}^d (s) \triangleq \frac{Z_s Z_e G_a}{Q Z_e G_a + (1 - Q) (Z_s + Z_e)}, \label{eq:Gss-d} \\
& \hat{G}_{se} (s) \triangleq \frac{(1 - Q) Z_s}{Q Z_e G_a + (1 - Q) (Z_s + Z_e)}. \label{eq:Gse-t}
\end{align}
Particularly, as $Q \rightarrow 1$ in \eqref{eq:Gss-d} and \eqref{eq:Gse-t}, we have $G_{ss}^d \rightarrow Z_s$ and $\hat{G}_{se} \rightarrow 0$, thereby imposing the reference model in Fig.~\ref{fig:model}(b). 

\subsection{Comparison to Direct P and PI Control} \label{sec:resemblance}

If we incorporate a first-order filter $Q(s) = \frac{\upalpha}{s + \upalpha}$ into the proposed torque controller in Sec.~\ref{sec:fs}, equations \eqref{eq:Gss-d} and \eqref{eq:Gse-t} can be expressed as follows:
\begin{align}
& G_{ss}^d (s) = \frac{Z_s Z_e G_a}{Z_s + Z_e + Z_e G_a \cdot \upalpha \cdot s ^ {-1}} \cdot \frac{s + \upalpha}{s}, \label{eq:Gss-d-1} \\
& \hat{G}_{se} (s) = \frac{Z_s}{Z_s + Z_e + Z_e G_a \cdot \upalpha \cdot s ^ {-1}}. \label{eq:Gse-t-1}
\end{align}
By having a small value for the pole $\upalpha$, we can neglect the term $Z_e G_a \cdot \upalpha \cdot s ^ {-1}$ in the denominators of \eqref{eq:Gss-d-1} and \eqref{eq:Gse-t-1}. Consequently, we obtain the approximations:
\begin{align}
& G_{ss}^d (s) \approx G_{sa}^d (s) \cdot \frac{s + \upalpha}{s}, & \mathrm{for \ small \ \upalpha}, \label{eq:Gss-d-2} \\
& \hat{G}_{se} (s) \approx G_{se} (s), & \mathrm{for \ small \ \upalpha}. \label{eq:Gse-t-2}
\end{align}
Notice that the term $\frac{s + \upalpha}{s}$ in \eqref{eq:Gss-d-2}, combined with the proportional gain $k$ outside of the disturbance observer (DOB) loop, constitutes a PI compensator $k \cdot \frac{s + \upalpha}{s}$, which is similar to the torque controller in \cite{senturk2018mri}.
However, in contrast to the direct P and PI control methods in \cite{wyeth2008demonstrating, sergi2015interaction, senturk2018mri}, our proposed control approach based on DOB yields several advantages. 

As the pole value $\upalpha$ increases from $0$ in \eqref{eq:Gss-d-1}, the term $\frac{s + \upalpha}{s}$ plays a crucial role in mitigating steady-state errors in the closed loop, presenting an improvement over a direct P controller in \cite{sergi2015interaction}. 
Moreover, with a continued increase in $\upalpha$, the influence of the term $Z_e G_a \cdot \upalpha \cdot s ^ {-1}$ in the denominators of \eqref{eq:Gss-d-1} and \eqref{eq:Gse-t-1} becomes more pronounced. 
This term, $Z_e G_a \cdot \upalpha \cdot s ^ {-1}$, drives $G_{ss}^d \rightarrow Z_s$ and $\hat{G}_{se} \rightarrow 0$, enhancing reference tracking performance and reducing sensitivity to external input $\uptau_e$ in the closed loop. 
These enhancements go beyond the capabilities of a direct PI controller. 

\subsection{Controller Parameter Tuning}

In Sec.~\ref{sec:resemblance}, we demonstrate that $G_{ss}^d \rightarrow Z_s$ and $\hat{G}_{se} \rightarrow 0$ as $\upalpha \rightarrow \infty$. 
Since $Z_s = \frac{k_s}{s}$ is the spring stiffness $k_s$ multiplied by an integrator $\frac{1}{s}$, it is straightforward to tune a proportional controller $\upomega_a^d = k \cdot (\uptau_s^d - \uptau_s)$ for the closed-loop torque control of our SEA. 

However, an infinitely high value of $\upalpha$ is impractical for several reasons. 
As $\upalpha \rightarrow \infty$, $Q (s) \rightarrow 1$, causing the inverse model $Z_s^{-1} = \frac{s}{k_s}$ in \eqref{eq:ve} not realizable in a digital system and increasing sensitivity to sensor noise from the spring deflection encoder. 
To mitigate sensitivity to sensor noise while using higher values of $\upalpha$, we employ $Q(s)$ as a second-order Butterworth filter $Q(s) = \frac{\upalpha^2}{s^2 + 2 \upzeta \upalpha s + \upalpha^2}$, where $\upzeta = \frac{\sqrt{2}}{2}$ ensures that $Q(s)$ has its cutoff frequency at $\upalpha$. 

Moreover, the $50 \ \mathrm{Hz}$ bandwidth of the motor velocity controller in $G_a(s)$ imposes constraints on the values of both the $Q(s)$ filter cutoff frequency $\upalpha$ and the proportional gain $k$. 
Adhering to this bandwidth limitation, the DOB-based controller is configured with $k = \frac{60}{k_s}$ and $\upalpha = 120 \ \mathrm{rad \cdot s^{-1}}$ ($\approx 19.1 \ \mathrm{Hz}$) throughout our experiments. 
This controller configuration is designed to provide a bandwidth of $60 \ \mathrm{rad \cdot s^{-1}}$ ($\approx 9.5 \ \mathrm{Hz}$) and a time constant of $\mathrm{t_c} = \frac{1}{60}$ sec for the closed-loop torque control of our SEA. 

\begin{figure} [!tbp]
\centering
\includegraphics[width=.99\linewidth]{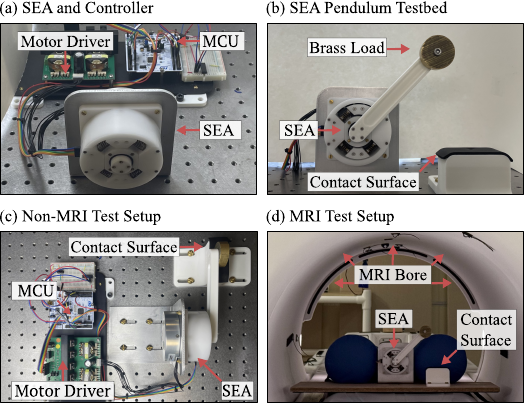}
\caption{
In (a), the SEA is controlled by a NUCLEO-F446RE MCU (STMicroelectronics, Geneva, Switzerland), which collects position data from encoders and sends velocity commands to the USM driver. 
In the non-MRI test setup (c), the SEA pendulum testbed (b) is positioned next to the USM driver and MCU. 
In the MRI test setup (d), only the SEA pendulum testbed is placed at the center of a $3$T MRI scanner bore, while the motor driver and MCU remain outside the MRI room, connected via 10-meter cables for power and control.
}
\label{fig:setup}
\end{figure}

\begin{figure*} [!tbp]
\centering
\includegraphics[width=.99\linewidth]{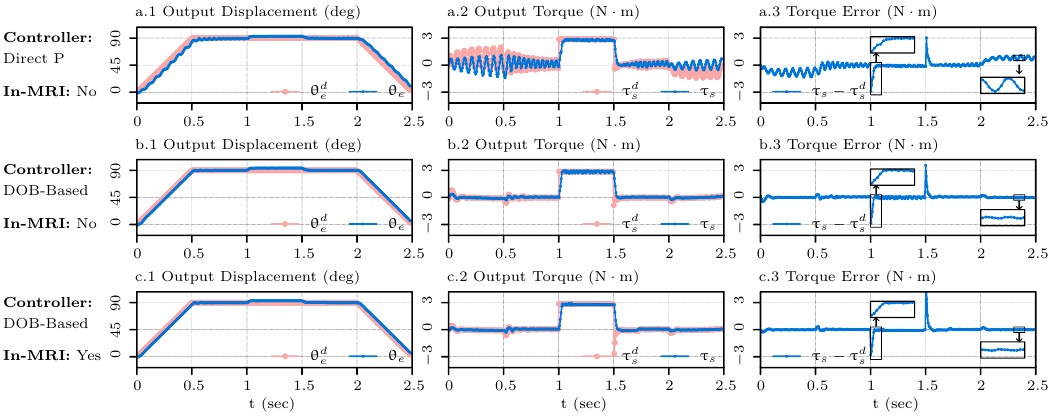}
\vspace{-10pt}
\caption{
Experimental results under different torque control and environmental settings are presented in distinct rows. 
The first column shows the SEA output displacement $\uptheta_e$, the second column represents the SEA output torque $\uptau_s$, and the third column displays the torque tracking error $\uptau_s - \uptau_s^d$. 
In a.3, b.3, and c.3, we highlight the transient responses for $\mathrm{t} = 1.0 \sim 1.1$ sec and $\uptau_s - \uptau_s^d = -3.3 \sim 0.3$ $\mathrm{N \cdot m}$, and the steady-state responses for $\mathrm{t} = 2.3 \sim 2.4$ sec and $\uptau_s - \uptau_s^d = -0.3 \sim 0.3$ $\mathrm{N \cdot m}$ (except for a.3, where $\uptau_s - \uptau_s^d = 7.7 \sim 8.3$ $\mathrm{N \cdot m}$). 
The small windows zoom in on the highlighted transient and steady-state response details, with arrows pointing from the highlights to the corresponding windows.
}
\label{fig:exp}
\end{figure*}

\section{Experimental Validation and Benchmarking} \label{sec:experiment}

In this section, we experimentally validate the SEA's capability to track torque references within and outside the MRI environment. 
For experiments outside the MRI environment, we present a performance evaluation comparing our DOB-based torque controller with the direct P torque controller in \cite{sergi2015interaction}.
Fig.~\ref{fig:setup} shows the hardware setup for the experimental validation of the SEA. 
The SEA is governed by a microcontroller unit (MCU), which acquires motor position $\uptheta_a$ and spring deflection $\uptheta_s$ measurements from encoders. 
Based on these measurements, the MCU sends motor velocity command $\upomega_a^d$ to the USM driver at $1000 \ \mathrm{Hz}$.

\subsection{Experimental Validation} \label{sec:T-A}

In the experiments, we connect the output gear of our SEA to a pendulum rod constructed from acrylonitrile butadiene styrene (ABS) via $3$D printing. 
As depicted in Fig.~\ref{fig:setup}(b), this pendulum rod supports a $125 \ \mathrm{g}$ brass load positioned $12.7$ cm from its center of rotation. 
This setup results in a moment of inertia of $m_e = 2.0 \times 10^{-3} \ \mathrm{kg \cdot m^2}$, which is slightly lower than the moment of inertia of a $1$ kg commercial TMS coil rotating about a wrist joint with a radius of $5$ cm.

In each experiment, the SEA swings the pendulum rod from $\uptheta_{e} = 0 \degree$ to $\uptheta_{e} = 90 \degree$ at $\mathrm{t} = 0 \sim 0.5$ sec. 
At $\mathrm{t} = 0.5 \sim 1$ sec, the brass load at the pendulum tries to contact a foam layer attached to a rigid object without exerting force, simulating contact with a human head in the \text{MRI}-guided brain stimulation application in Fig.~\ref{fig:tms}. 
At $\mathrm{t} = 1 \sim 1.5$ sec, the SEA exerts a $3$ $\mathrm{N \cdot m}$ torque to press against the contact surface. 
Finally, the SEA releases the additional torque at $\mathrm{t} = 1.5 \sim 2.0$ sec and swings the pendulum rod back to $\uptheta_{e} = 0 \degree$ at $\mathrm{t} = 2 \sim 2.5$ sec. 
For the experiment in the MRI environment, this setup is positioned to the center of the bore in a $3$T GE MR750W scanner (GE Healthcare, Waukesha, WI, USA), as depicted in Fig.~\ref{fig:setup}(d).

To track the desired output position $\uptheta_e^d$ of the SEA in the experiments, we implement an impedance controller 
\begin{equation}
\uptau_s^d = k_v \cdot (\uptheta_e^d - \uptheta_e) + b_v \cdot (\upomega_e^d - \hat{\upomega}_e), \label{eq:ksv}
\end{equation}
where $k_v$ is a virtual spring stiffness, $b_v$ is a virtual damper coefficient, $\upomega_e^d$ is set to be zero, and $\hat{\upomega}_e$ is the output velocity estimate introduced in \eqref{eq:ve}. 
We set $k_v = 0.5 \cdot k_s$ and $b_v = 1.5 \times 10 ^ {-2} \cdot k_s$, allowing the SEA to respond to the external force and dynamics with a stiffness reduced to half of its original value $k_s$ and to swing the pendulum rod with a damping ratio of $0.9$.

In Fig.~\ref{fig:exp}(a)-(b), we present experimental results outside the MRI environment, comparing the direct P controller from \cite[Fig.~3(c)]{sergi2015interaction} with our DOB-based controller. 
The proportional gains for both controllers are set to $k = \frac{60}{k_s}$.  
In Fig.~\ref{fig:exp}(b), the time response of the output torque tracking error, $\uptau_s - \uptau_s^d$, with the DOB-based controller exhibits a $5\%$ settling time of approximately $0.05$ sec. 
As the $5\%$ settling time corresponds to three times the time constant of the closed-loop SEA, the experimentally observed time constant aligns with the desired value of $\mathrm{t_c} = \frac{1}{60}$ sec.
Except for the response during $\mathrm{t} = 1.0 \sim 1.5$ sec, the DOB-based controller maintains a steady-state error of $\uptau_s - \uptau_s^d$ within $\pm 0.08$ $\mathrm{N \cdot m}$, equivalent to $2.5\%$ of the maximum output torque.  
The increased steady-state error observed at $\mathrm{t} = 1.0 \sim 1.5$ sec can be attributed to the virtual damper coefficient, $b_v$, being tuned for the pendulum swing phase. 
This tuning renders the SEA under-damped when pressing against the high impedance of a rigid object.

In contrast, the direct P controller destabilizes the SEA when not in contact with a rigid object and introduces a steady-state torque error bias of $0.80$ $\mathrm{N \cdot m}$, equivalent to $25\%$ of the maximum output torque, as shown in Fig.~\ref{fig:exp}(a).

Fig.~\ref{fig:exp}(c) presents the experimental results for the DOB-based controller in the MRI environment. 
The response retains a $5\%$ settling time of approximately $0.05$ sec and a steady-state error of $\uptau_s - \uptau_s^d$ within $\pm 0.08$ $\mathrm{N \cdot m}$. 
Notably, this steady-state error bound is also observed during $\mathrm{t} = 1.0 \sim 1.5$ sec, indicating improved performance compared to the results in Fig.~\ref{fig:exp}(b). 
This improvement can be explained by the Lorentz force in the MRI environment, which causes the metal spring to behave like a viscoelastic spring, mitigating the under-damped behavior during rigid contact.    
Apart from this improvement, the consistent time responses in Fig.~\ref{fig:exp}(b)-(c) confirm our SEA's suitability for MRI applications.

\subsection{Benchmarking Analysis} \label{sec:benchmarking}

When implementing the virtual impedance controller in \eqref{eq:ksv} in a digital system, it can be approximated as a lead controller $k_v \cdot \frac{p}{z} \cdot \frac{s + z}{s + p}$, where $z = \frac{k_v}{b_v}$ and $p$ is a large value for numerical differentiation. 
In \cite[Eq.~(13)]{sergi2015stability}, the coupled stability of a velocity-sourced SEA with external inertia $m_e$ is analyzed, demonstrating that a direct P controller in the inner torque loop and a lead controller in the outer impedance loop require a minimum value of $m_e$ for stability. 
This explains why the direct P controller is unstable when swinging the pendulum alone but stabilizes when pressing a rigid object. 
While the DOB in our controller is designed for the inner torque loop, it compensates for low inertia during pendulum swings, ensuring stability in the outer impedance loop.

Assuming $G_a (s) \approx 1$ at low frequencies, the closed-loop transfer function for the direct P control method in \cite{sergi2015interaction} is $G_\mathrm{CL}^\mathrm{P} (s) = \frac{k \cdot G_{sa} (s)}{1 + k \cdot G_{sa} (s)}$, where $G_{sa} (s)$ is defined in \eqref{eq:Gsa}. 
When $Z_e (s)$ is small, it simplifies to $G_\mathrm{CL}^\mathrm{P} (s) \approx \frac{k \cdot Z_e (s)}{1 + k \cdot Z_e (s)}$, resulting in a closed-loop gain significantly below $1$. 
In the outcomes of pendulum swing phase, the direct P controller exhibits inadequate steady-state performance under low external impedance scenarios. 
When $Z_e (s)$ is large, $G_{sa} (s) \approx Z_{s} (s) = \frac{k_s}{s}$, resulting in a closed-loop gain $G_\mathrm{CL}^\mathrm{P} (s) \approx 1$ in the steady state. 
Consequently, the performance improves considerably under high external impedance conditions, as shown at $\mathrm{t} = 1 \sim 1.5$ sec.

In \cite{wyeth2008demonstrating, senturk2018mri}, velocity-sourced SEAs incorporate a PI compensator to mitigate steady-state tracking errors. 
In contrast to the 1st-order PI controller employed in \cite{senturk2018mri}, the SEA system presented in \cite{wyeth2008demonstrating} utilizes a 2nd-order PI controller by cascading two 1st-order PI controllers. 
The design of this 2nd-order PI controller strategically positions its two zeros to cancel the resonant poles of $G_{sa}(s)$ in \eqref{eq:Gsa}. 
However, without prior knowledge of $Z_e(s)$, both types of PI compensators can introduce stability issues in the closed-loop SEA system.
Although a direct comparison with the MARC and ARC methods from \cite{losey2016time, calanca2018understanding} is not performed, our results demonstrate that the DOB-based controller achieves stable and robust performance under the contact-switching scenarios.

\section{Discussion and Future Work} \label{sec:discussion}

In this section, we further discuss the results of our 1-DOF SEA evaluation, highlighting its mechanical design, MRI compatibility, and control performance. 
We also outline the potential challenges of scaling this approach to a multi-DOF SEA robot for MRI applications and indicate how our findings may offer useful insight for addressing these challenges in future developments.

\subsection{Failure Modes and Mitigation} \label{sec:failure}

The TFSEA offers a compact structure well suited for MRI environments, but this advantage comes with a trade-off in reduced output force sensing accuracy due to mechanical friction and backlash in the drivetrain. 
As shown in the appendix, the planetary gearbox introduces a peak-to-peak frictional torque of $0.15$ $\mathrm{N \cdot m}$ (2.3\% of the total spring torque range) and a peak-to-peak backlash of $1.38\degree$ (in the SEA output frame). 
While drivetrain losses do not dominate spring deflection in our prototype, offline calibration may still be applied to compensate for friction and backlash.
As reviewed in \cite{huang2019intelligent}, both model-based methods and model-free techniques, such as neural networks and fuzzy logic, offer promising solutions.

For the torque control of our SEA, extremely low output impedance presents another potential failure mode. 
Theoretically, the effectiveness of SEA torque control diminishes as the output impedance approaches zero. 
In near-free output conditions, the controllability of spring deflection regulation decreases significantly. 
This issue is further exacerbated by the limited power density of the USM, which imposes tight constraints on input velocity. 
Although other velocity-sourced motors with MRI compatibility and higher power density \cite{gunderman2023modeling, pan2024cornerstone, gassert2006mri} can also be integrated into our SEA module, USMs offer superior velocity control accuracy and bandwidth. 
To mitigate this controllability limitation, a minimum allowable output impedance should be incorporated into the controller design. 

\subsection{Additional MR Imaging Insights} \label{sec:insights}

The MR imaging test in Sec.~\ref{sec:image} evaluates $\mathrm{B_0}$ field distortion caused by the SEA materials and the SNR change with the SEA powered off but connected via a shielded and filtered cable from the control room. 
These results confirm that the imaging artifacts and signal noise associated with the metal spring and fasteners, the USM, and the encoder remain within acceptable limits. 
At the same time, the necessity of including these components in our SEA module is also apparent. 
The metal spring and its encoder are essential for generating and measuring the SEA spring torque, the USM provides precise and high-bandwidth motor velocity control inside the MRI scanner, and the metal fasteners supply the structural strength that plastic alternatives cannot achieve.

Although the imaging tests exclude dynamic motion and active electronics, the setup captures the essential operating condition in which the robot remains stationary during MRI scans and moves only between scans to reposition and carry out medical tasks.
Due to the non-backdrivability of the USM when powered off, the robot stays fixed in place.
This operating scheme helps avoid potential imaging artifacts from actuator motion and electromagnetic interference from powered components.

\subsection{Implications for Multi-DOF SEA Robots in MRI Scanners} \label{sec:multi-dof}

While the prototype demonstrates the promise of the compact rotary SEA module for MRI-compatible robotics, this paper does not address the full drivetrain that transmits SEA torque to robot joints. 
Designing an MRI-compatible drivetrain that is both compact and sufficiently stiff to limit end-effector backlash remains challenging. 
Nonetheless, the SEA’s compact form factor ($80 \times 80 \times 66$ mm) conserves space in the MRI bore and provides flexibility for system integration. 
The module also supports future upgrades, including customized MR-safe springs \cite{sergi2015interaction} and alternative coaxial gearboxes such as cycloidal drives \cite{kim2021compact}, which may increase torque for elbow and shoulder actuation in the MRI-guided brain stimulation application in Fig.~\ref{fig:tms} and improve energy efficiency \cite{nieto2019minimizing} and torque bandwidth \cite{shu2024impact}.

Controlling SEA-based multi-DOF robots introduces internal compliance dynamics that complicate coordination relative to rigid systems. 
A common solution is a cascade structure that separates joint-space torque regulation from task-space impedance control \cite{thomas2021formulating}. 
As described in Sec.~\ref{sec:experiment}, our controller follows this paradigm, with an inner DOB-based torque loop and an outer impedance loop. 
A key requirement for robust whole-body control of SEA robots is high-fidelity joint-level torque tracking \cite{paine2013design}, which can degrade due to inertia variations from robot configuration changes. 
Our DOB-based controller is designed to compensate for these variations, enabling consistent torque tracking across postures.

In Sec.~\ref{sec:experiment}, we evaluate our DOB-based controller under three distinct contact conditions: no contact, contact without applied force, and contact with applied force. 
While the controller demonstrates robust performance across all conditions in our 1-DOF experimental setup, human-robot contact scenarios in multi-DOF clinical systems are inherently more complex.
Specifically, the simplified point contact considered in this paper can become a line or surface contact involving multiple degrees of freedom, variable orientations, and potential misalignment between the TMS device and the patient’s scalp. 
Effectively managing contact conditions across different orientations and constraints in multi-DOF robotic systems remains an important direction for future research.

\section{Conclusion}

In this paper, we present a compact and efficient rotary SEA module that is MRI-compatible and powered by velocity-sourced ultrasonic motors. 
Using a DOB-based controller, we achieve robust torque control across different output impedance and contact conditions. 
Experiments conducted in both 3-Tesla MRI and non-MRI environments demonstrate consistent performance, with a $5\%$ settling time of 0.05 seconds and a steady-state torque error within $2.5\%$ of the SEA's maximum output torque.

\section*{Appendix}
\addcontentsline{toc}{section}{Appendix}

\subsection{Gearbox Friction and Backlash}

\begin{figure} [!tbp]
\centering
\includegraphics[width=0.99\linewidth]{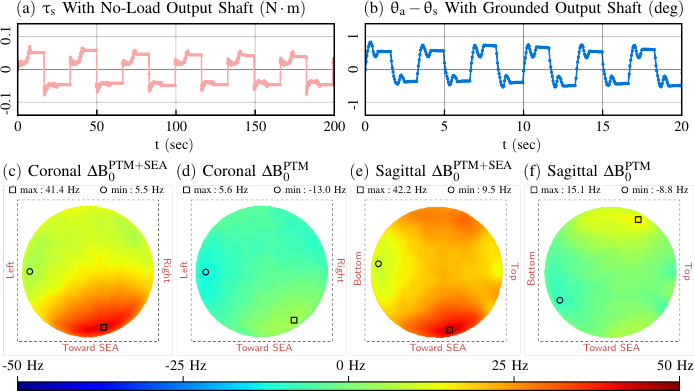}
\caption{
Gearbox friction is measured by slowly rotating the motor with an unloaded output shaft, with $\uptau_s$ shown in (a). Backlash is measured by fixing the output and commanding small motor displacements, with $\uptheta_a - \uptheta_s$ shown in (b).
For MR imaging, the $\mathrm{B_0}$ deviation with the SEA, $\Delta \mathrm{B_0}^\mathsf{PTM+SEA}$, is shown in the coronal (c) and sagittal (e) planes, compared to the baseline $\Delta \mathrm{B_0}^\mathsf{PTM}$ without the SEA in (d) and (f).
}
\label{fig:supplement}
\end{figure}

To characterize the gearbox friction and backlash of our SEA prototype, we perform two tests. 
For friction measurement, the output shaft is left unloaded while the motor slowly rotates in both directions. 
The frictional torque is approximated by the spring torque $\uptau_s$, as the spring should remain at its origin position in the absence of friction. 
For backlash measurement, the output shaft is fixed, and the motor is commanded to oscillate over a small angular range. 
The backlash is estimated as the difference between the motor displacement $\uptheta_a$ and the spring deflection $\uptheta_s$, which would be equal in an ideal system. 
Fig.~\ref{fig:supplement}(a)-(b) shows the time plots of $\uptau_s$ and $\uptheta_a - \uptheta_s$, indicating the peak-to-peak values of gearbox friction and backlash.

\subsection{MR Imaging Test Settings}

To evaluate artifacts induced by the SEA, $\mathrm{B_0}$ mapping and SNR measurements are performed using gradient echo (GRE) sequences. 
A 2D multi-slice two-echo GRE sequence is used for $\mathrm{B_0}$ mapping with the following parameters: FOV = $18 \times 18$ cm, matrix size = $256 \times 256$, TR = 500 ms, TE1 = 4.4 ms, TE2 = 9.4 ms, flip angle = 20$^\circ$, and slice thickness = 3 mm. 
Two setups are assessed in this test: the baseline setup without the presence of the SEA and the setup with the unpowered SEA. 
The test sequence is as follows:
(i) Shimming is performed on the phantom region using the baseline setup. 
(ii) $\mathrm{B_0}$ map is acquired for the baseline setup. 
(iii) $\mathrm{B_0}$ map is acquired for the setup shown in Fig.~\ref{fig:image}(b), without re-shimming. 
Following this sequence, we obtain $\mathrm{B_0}$ map images for both the coronal and sagittal planes.

In Fig.~\ref{fig:supplement}, we present the $\mathrm{B_0}$ field deviations, $\Delta \mathrm{B_0}^\mathsf{PTM}$ and $\Delta \mathrm{B_0}^\mathsf{PTM+SEA}$, for the baseline setup and the setup shown in Fig.~\ref{fig:image}(b). 
In the notations of $\Delta \mathrm{B_0}$, the superscripts $\mathsf{PTM}$ and $\mathsf{SEA}$ are used to denote the phantom and the SEA, respectively. 
To isolate the SEA-induced $\mathrm{B_0}$ field deviation, $\Delta \mathrm{B_0}^\mathsf{SEA}$ is computed as $\Delta \mathrm{B_0}^\mathsf{SEA} = \Delta \mathrm{B_0}^\mathsf{PTM+SEA} - \Delta \mathrm{B_0}^\mathsf{PTM}$, eliminating the baseline setup's $\mathrm{B_0}$ inhomogeneity. 
In Fig.~\ref{fig:image}(c)-(d), we show the $\mathrm{B_0}$ field deviation of $\Delta \mathrm{B_0} = \Delta \mathrm{B_0}^\mathsf{SEA}$.

For the SNR measurement, we use a multi-slice 2D GRE sequence with parameters: slice thickness = $5$ mm, inter slice spacing = $5$ mm, a total of 16 slices covering the phantom, FOV = $20 \times 20$ cm, matrix size = $256 \times 256$, TR = 300 ms, TE = 10 ms, and flip angle = 60$^\circ$. 
The average SNR is computed across all pixels within the phantom volume. 
A second acquisition with identical parameters, but with the transmit RF pulses disabled, was performed immediately afterward to obtain a noise-only image for estimating the noise standard deviation.

\balance

\bibliographystyle{IEEEtran}

\end{document}